\documentclass{ieeeaccess}
\usepackage{cite}
\usepackage{amsmath,amssymb,amsfonts}
\usepackage{algorithmic}
\usepackage{graphicx}
\usepackage{textcomp}
\usepackage{subfigure}

\usepackage{CJK}
\usepackage{soul}
\usepackage{color}

\setulcolor{red} 
\setstcolor{yellow} 
\sethlcolor{red} 

\def\BibTeX{{\rm B\kern-.05em{\sc i\kern-.025em b}\kern-.08em
    T\kern-.1667em\lower.7ex\hbox{E}\kern-.125emX}}
    
\graphicspath{{figures/}{figures/NYU/}{figures/MSRA/}{figures/ICVL/}} 
    
\begin{document}
\history{Date of publication xxxx 00, 0000, date of current version xxxx 00, 0000.}
\doi{10.1109/ACCESS.2017.DOI}

\title{HMTNet: 3D Hand Pose Estimation from Single Depth Image Based on Hand Morphological Topology}
\author{
\uppercase{Weiguo Zhou}\authorrefmark{1}, \IEEEmembership{Student Member, IEEE},
\uppercase{Xin Jiang}\authorrefmark{1}, \IEEEmembership{Member, IEEE}, 
\uppercase{Chen Chen}\authorrefmark{1}, 
\uppercase{Sijia Mei}\authorrefmark{1}, 
\uppercase{and Yun-Hui Liu}\authorrefmark{2}, \IEEEmembership{Fellow, IEEE}}
\address[1]{Department of Mechanical Engineering and Automation, Harbin Institute of Technology, Shenzhen, China (e-mail:weiguochow@gmail.com)}
\address[2]{T Robotics Institute, The Chinese University of
Hong Kong, Hong Kong, China}
\tfootnote{This work was supported in part by Shenzhen Pea-cock Plan Team under Grant KQTD20140630150243062 and in part by Shenzhen and Hong Kong Joint Innovation Project under Grant SGLH20161209145252406.}

\markboth
{Zhou \headeretal: HMTNet: 3D Hand Pose Estimation from Single Depth Image Based on Hand  Morphological Topology}
{Zhou \headeretal: HMTNet: 3D Hand Pose Estimation from Single Depth Image Based on Hand  Morphological Topology}

\corresp{Corresponding author: Xin Jiang (e-mail:x.jiang@ieee.org).}

\begin{abstract}
Thanks to the rapid development of CNNs and depth sensors, great progress has been made in 3D hand pose estimation. Nevertheless, it is still far from being solved for its cluttered circumstance and severe self-occlusion of hand. In this paper, we propose a method that takes advantage of human hand morphological topology (HMT) structure to improve the pose estimation performance. The main contributions of our work can be listed as below. Firstly, in order to extract more powerful features, we concatenate original and last layer of initial feature extraction module to preserve hand information better. Next, regression module inspired from hand morphological topology is proposed. In this submodule,
we design a tree-like network structure according to hand joints distribution to make use of high order dependency of hand joints. Lastly, we conducted sufficient ablation experiments to verify our proposed method on each dataset. Experimental results on three popular hand pose dataset show superior performance of our method compared with the state-of-the-art methods. On ICVL and NYU dataset, our method outperforms great improvement over 2D state-of-the-art methods. On MSRA dataset, our method achieves comparable accuracy with the state-of-the-art methods. To summarize, our method is the most efficient method which can run at $220.7$ fps on a single GPU compared with approximate accurate methods at present. The code will be available at\footnote{https://github.com/weiguochow/HMTNet}.
\end{abstract}

\begin{keywords}
3D hand pose estimation, concatenated feature, hand morphological topology, single depth image.
\end{keywords}

\titlepgskip=-15pt

\maketitle

\section{Introduction}
\PARstart{H}{and} pose estimation attracts plenty of attention in various applications such as robotics, human-robot interaction, virtual reality/augmented reality, etc. Benefit from the rapid development of depth sensors, CNNs, and computational ability of GPU, 3D hand pose estimation using CNNs has shown a great advantage over traditional methods. Recently, despite a rapid growing number of publications \cite{khamis2015learning,joseph2016fits,sridhar2015fast,sharp2015accurate} on hand pose estimation, but it is still a challenging problem far from solved for following difficulties, complex background, high degree of freedom of hand, small hand area of image, low resolution and noise of depth sensors, and self-occlusion of hand. 

In these literatures, 3D hand pose estimation methods fall into three categories. The first one is model based (generative) method which finds the joints position to minimize the cost energy \cite{khamis2015learning,joseph2016fits,sridhar2015fast,sharp2015accurate} between the model and estimated joints. Another one is data driven (discriminative) method \cite{moon2017v2v,oberweger2017deepprior++,oberweger2015hands,madadi2017end,guo2017towards,ge20173d,deng2017hand3d,chen2017pose} which derives the positions based on the experience gained from dataset directly. Then the last category is hybrid method \cite{wan2017crossing,zhou2016model} which takes advantage of hand model based method and data driven method. With the success of deep learning technique, convolution neural network based 3D hand pose estimation has became the leading method. We will review the related literatures on 3D hand pose estimation based on deep learning especially using single depth image.

Thompson\cite{tompson2014realtime} is the pioneer who used deep learning to extract joint positions. Firstly, the depth map is calculated by using a convolution network, and then a subdivided heat map is generated. Then they used the inverse dynamics method to combine the heat map and features to derive the hand posture. It was found from that it can realize $40$ fps using only CPU. But it can only obtain two-dimensional positions of the hand joints, so it needs to map three-dimensional depth map to obtain the coordinate values of the joints.

Oberweger \cite{oberweger2015training} proposed to use feedback loops to correct prediction errors. The depth map is used to estimate the 3D pose of gestures. The pose error can be corrected by a convolution neural network. The feedback is also a deep network that can be optimized through training data.

There is a trend that more and more attention paid on the hand kinematics or hand model to take full advantage of the prior information of hand. In this method \cite{zhou2016model}\cite{madadi2017end}\cite{chen2017pose}\cite{zhou2018hbe}\cite{du2019crossinfonet}\cite{madadi2017end}, hand physics structure and kinematic are involved in many situations. Hand model kinematics and geometric constraints are considered.

Our main contribution lies in two aspects, the first one is a more robust feature extraction module which adopt the first and last layer of the initial feature extraction module and the second one is hand pose regression module inspired from hand morphological topology structure. Main contributions of this paper can be summarized as follows:

\begin{itemize}
	\item A richer initial feature extraction module was designed which concatenated original and last layer of this module. This concatenated feature can be better represent the input depth map of hand.
	\item To take full advantage of hand morphological topology, we design a tree-like branch structure network according to hand joints distribution. This can be seen as the different range of motion via the distribution order belonging to each finger.
	\item We take a sufficient ablation study of our proposed architecture and obtain impressive performance on popular hand pose dataset. The experimental results show that our method achieved superior performance over current excellent performance in terms of both accuracy and efficiency.
\end{itemize}

In the following, structure of this paper is given. Section II introduces the related works on 3D hand pose estimation especially using hand kinematic or morphology information. Our proposed method is illustrated in detail in Section III. The following section is the experimental study including ablation study and the comparison with the state-of-the-art performance. Conclusions and future work are also given in Section V.

\section{Related Works}
Hand pose estimation has attracted a lot of attention due to its widely application. Our work focuses on depth image based hand pose estimation, so we will review depth map based 3D hand pose estimation methods especially related to hand kinematic and morphological topology structure. These methods can be categorised into joining loss constraints \cite{zhou2016model}\cite{madadi2017end} and branch-like network architecture \cite{chen2017pose}\cite{zhou2018hbe}\cite{du2019crossinfonet}\cite{madadi2017end} in these publications. In the following, we will review the related works using hand model or hand kinematic constraints.

\subsection{Hand Kinematic Constraints}

Deep-Prior \cite{oberweger2015hands} proposed by Oberweger is widely used recently. The highlight of this paper is the usage of a low dimensional bottleneck in the last fully-connected layer. This layer forces the network to learn a low dimensional feature which is similar with embedded physical constraints on network architecture.

Model base network architecture for hand pose estimation \cite{zhou2016model} is proposed to fully exploit the hand model prior information. Specially, the authors designed an efficient, parameter-free and differential layer that realized non-linear forward kinematics function. In such architecture, forward kinematic process of hand is integrated to realize the hand kinematic constraints. By this method, infeasible physical hand pose position was limited and then superior performance was achieved. However, there is a limitation on generalization since the hand shape and length are fixed constant. Furthermore, different from \cite{Zhou_2017_ICCV}, whose network is trained in three stages. In this work, an end-to-end network is proposed.

Loss function is also used to incorporate physical constraints of the infeasible position. In this work\cite{madadi2017end}, the authors took four loss terms (these are, global loss, local loss, appearance loss, and dynamics loss, respectively) into consideration. Joints must locate inside hand area, that is, depth value higher than hand surface. Next, the authors assumed all joints belonging to each finger should be collinear or coplanar.

\subsection{Branch-like Network Architecture}
In \cite{madadi2017end}, hierarchical tree-like structured CNN was proposed. Firstly, local pose features are extracted by predefined set of hand joints and then fuse these local features to learn dependencies on joints. Besides, physical constraints are joined to the loss terms to avoid infeasible joint locations.

Pose-REN \cite{chen2017pose} extracts the feature map region based on the initial estimated pose. This method addresses the problem that treating the feature regions equally. Features are partitioned according to hand topology, one finger has the same feature region. It ensemble the feature map according to the topology of hand joints.

In \cite{zhou2018hbe}, three branches network named hand branch ensemble network (HBE) is proposed. In this architecture, three branches represent three parts of a hand, these are, thumb, index and other fingers, respectively. The idea involved in this architecture is inspired from the understanding to the different usage of fingers in daily manipulation. Besides this, a low-dimensional embedding layer is also adopted to join the hand kinematic constraints.

\Figure[t!](topskip=0pt, botskip=0pt, midskip=0pt)[width=7.0in]{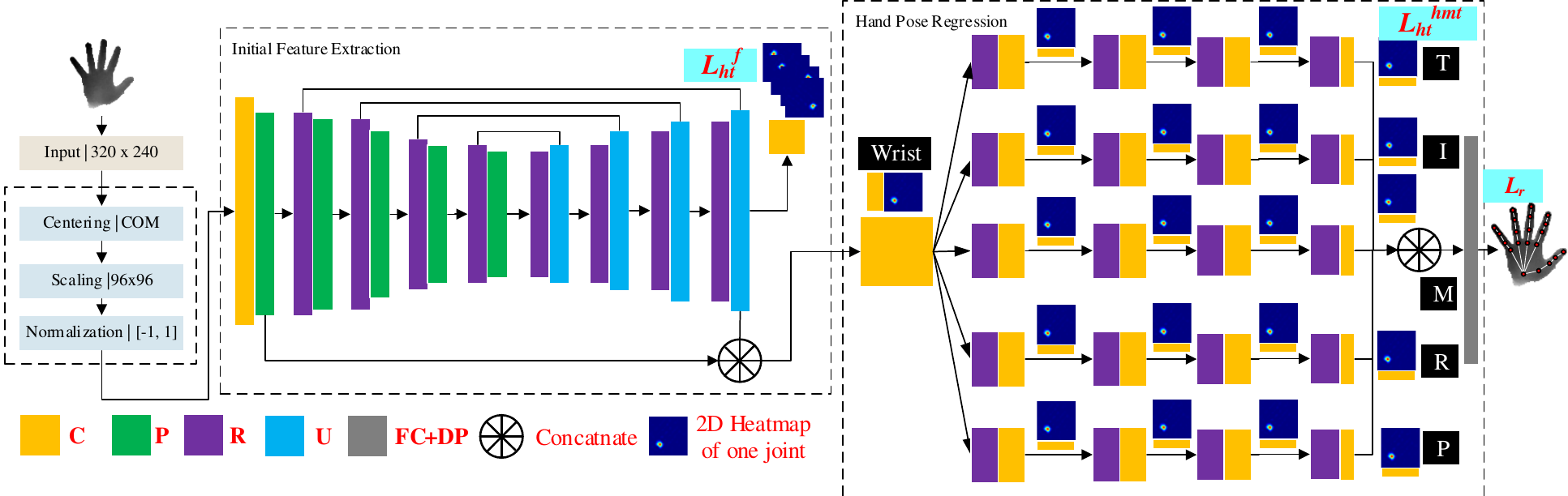}
{proposed 3D hand pose estimation network with hand morphological topology structure. In the hand pose regression module, we take MSRA dataset as example. \textbf{C} denotes convolution layers, \textbf{P} means pooling layer and \textbf{R} stands for residual module, \textbf{U} is the upsampling module, and \textbf{FC+DP} represents fully connected and dropout layer module, respectively. \textbf{T} stands for thumb finger, \textbf{I} denotes index finger, \textbf{M} means middle finger, \textbf{R} is the ring finger and \textbf{P} is pinky finger. The total loss of this architecture contains three parts, these are, $\mathcal{L}_{ht}^{f}$, $\mathcal{L}_{ht}^{hmt}$, and $\mathcal{L}_{r}^{f}$.
\label{fig_HMT}}

CrossInfoNet \cite{du2019crossinfonet} inspired from the multi-task mechanism separate the hand pose estimation task into palm pose estimation subtask and finger pose estimation subtask, respectively. Based on this structure, the two branches can share information with each other. The total loss contains three parts, these are, guided heatmap loss, feature refinement loss, and regression loss, respectively.

\Figure[t!](topskip=0pt, botskip=0pt, midskip=0pt)[width=2.0in]{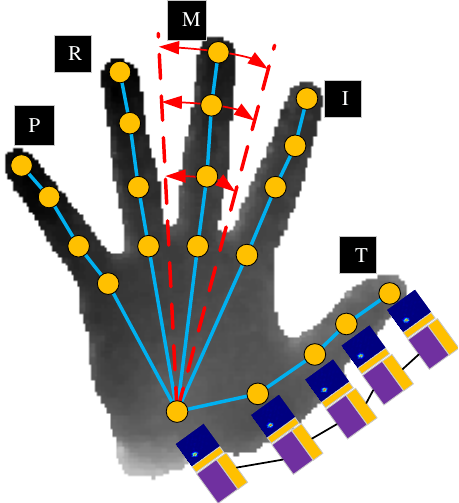}
{Hand morphology topology used on our experiment. The location of far-end joints are depend on the near-end joints relative to root joint. Then in this experiment, we design hand morphological topology-like network to imitate this structure to learn the dependency among these joints. The latter generated heatmap, that is the joint location, depended on the former joints.
\label{fig_hand}}

\section{Methodology}

In this section, we present our proposed method in details. In this paper, our aim is to estimate 3D hand joint locations $\mathbf{\Phi }=\left \{ \phi_{i} \right \}, i=1, ...,$ J with $\phi_{i}=\left ( x_{i}, y_{i}, z_{i} \right)$ from a single depth image $\mathbf{D}$ $\in$ $\mathbb{R}^{640 \times 480}$. Firstly, We introduce training data preprocessed and data argument, and then we describe our network architecture which consists of two stages, initial feature extraction module and hand pose regression module. 

Overall architecture of this network is shown in Fig.\ref{fig_HMT}. In general, the network can be split into two main modules. The first is initial feature extracted module, and the other is hand pose regression module.

\subsection{Dataset Preprocess and Training Dataset Augmentation}

In data preprocess stage, we assume the hand is closest to the camera. Then we can detect the hand according to the depth value. Then, the center location of hand is obtained to derive the 3D position of center of hand according the intrinsic parameters. A 3D cube with size ($250$ mm, $250$ mm, $250$ mm) located at the hand center of mass (COM) is cropped. Then the cube is mapping into depth map according to the intrinsic parameters, then depth map is resized to what the network requires which is $96 \times 96$. In our work, we set adapted parameter to connect the 2D heatmaps and depth regression. In actual experiment, the depth value is normalized to range $[-1, 1]$ which is fed into the following network architecture to make the network adjust according to the different distance from hand to camera.

The online training data augmentation via translation, rotation and scaling was done using the same way as \cite{oberweger2017deepprior++}. In this augmentation, scaling ,translation, and rotation are used. By means of data augmentation, the network architecture can be more robust to environment changing.

\subsection{Initial Feature Extraction Module}
In the first module, we used the residual module as the backbone architecture for its success on image classification task. Besides this, we also concatenate all layers of network to obtain richer feature. Specially, we set heatmap guided on all hand joints independently from hand root (wrist) joint to fingertip. In the ablation study, the results show that our extracted module is superior than other methods.
Inspired by recent works on pose estimation, we want to obtain a richer initial extracted feature. As in CrossInfoNet \cite{du2019crossinfonet}, a heat map guided feature extraction module is also used. In this paper, we will use this module as the baseline of initial feature in ablation study. We concatenate the original feature and last feature to obtain a more nutritious feature for the next hand pose regression module. In this, we think concatenated feature contain the original and high-dimensional feature.

\subsection{Hand Pose Regression Module based on Hand Morphological Topology}

Previous tree-like branch methods treat the joints belonging each finger equally, which is not optimal enough to fully incorporate the spatial information of hand structure and obtain highly representative features. Not only different fingers, but also different joints belonging the same finger are considered on our method.

As shown in Fig.\ref{fig_hand}, the range of movement of hand joints is depended on the order of each hand. Far-end joints are depend on near-end joints which relative to the root joint. Then, we not only split the hand into branch according to finger. Moreover, we split the hand according the distribution of hand joints not only finger by finger but also joint by joint.

\Figure[t!](topskip=0pt, botskip=0pt, midskip=0pt){base_base}
{Baseline architecture for ablation study. A single channel network with same initial feature extraction except no concatenating the large feature channel. 
In this figure, \textbf{C} denotes convolution layers, \textbf{P} means pooling layer, \textbf{R} stands for residual module, \textbf{U} stands for upsampling module, and \textbf{FC+DP} means fully connected and dropout layer, respectively. In our work, heatmap is generated by the gaussian function distribution centering on the location of hand joint. The total loss of this architecture contains two parts, these are, $\mathcal{L}_{ht}^{f}$, and $\mathcal{L}_{r}^{f}$.\label{fig_base_base}}

\subsection{Loss Functions}

The total loss of this network architecture contains three parts, which is heatmap guided loss, joint regression loss, and joints heatmap loss. They are defined in following equations, respectively. The total loss can be expressed in (\ref{loss_total}).
\begin{equation}\label{loss_total}
\begin{split}
{\mathcal{L}_{total}} = \lambda_{f}{\mathcal{L}_{ht}^{f}} + {\lambda _{r}}{\mathcal{L}_{r}} + {\lambda _{hmt}}{\mathcal{L}_{ht}^{hmt}} + {\lambda _{w}}{\mathcal{R}(w)},
\end{split}
\end{equation}
where ${\lambda _{f}}$, ${\lambda _{r}}$, ${\lambda _{hmt}}$ are balance factor to weight the three loss terms, respectively. We set ${\lambda _{f}}=0.005$, ${\lambda _{hmt}} = 0.005$, ${\lambda _{r}} = 0.05$, and ${\lambda _{w}} = 1$ in our experiment. $\mathcal{L}_{ht}^{f}$ is the heatmap guided loss which is to drive a more accurate initial extracted feature. $\mathcal{L}_{r}$ is the regression loss of all joints. $\mathcal{L}_{ht}^{hmt}$ is the tree-branch part heatmap loss of all joints independently. $\mathcal{R}(w)$ is the weight regularization term which is $\frac{1}{2}\sum w^{2}$. In the following, each component is explained in detail.

\subsubsection{Feature Heatmap Loss}
In this module, heatmap guided loss is used to obtain a more accurate feature map. 

\begin{equation}\label{eq:ht_loss}
{\mathcal{L}_{ht}^{f}} =  {\sum\limits_{i=1}^{J} \sum\limits_{u}^{w} \sum\limits_{v}^{h}{{{\left \| {\mathbf{H}_{i}^{f(est)}(u,v) - \mathbf{H}_{i}^{f(gt)}(u,v)} \right \|}^2}} },
\end{equation}
where $\mathbf{H}_{i}^{f(est)} \in {\mathbb{R}^{w \times h}}$ is the heatmps corresponding to estimated joint coordinates and $\mathbf{H}_{i}^{f(gt)} \in {\mathbb{R}^{w \times h}}$ denotes heatmaps corresponding to ground truth joint coordinates in the initial feature extraction module. In this work, $24 \times 24$ feature map is used to supervise the heatmap loss term, that is $w,h$ equals to $24$.

\subsubsection{Joints Regression Loss}
\begin{equation}\label{eq:reg_loss}
{\mathcal{L}_{reg}} =  {\sum\limits_{i=1}^{J} {\left \| \phi_{i}^{est} - \phi_{i}^{gt} \right \|}^2_2},
\end{equation}
where $\phi_{i}^{est} \in \mathbb{R}^3$ and $\phi_{i}^{gt} \in \mathbb{R}^3$ represent the $i$th estimated and ground-truth joints locations, respectively.

\subsubsection{Joints Heatmap Loss}
In this module, we use every point heatmap supervised

\begin{equation}\label{eq:joint_ht_loss}
{\mathcal{L}_{ht}^{hmt}} =  {\sum\limits_{i=1}^{J} \sum\limits_{u}^{w} \sum\limits_{v}^{h} {{{\left \| {\mathbf{H}_{i}^{hmt(est)}(u,v) - \mathbf{H}_{i}^{hmt(gt)}(u,v) } \right \| }^2}} } ,
\end{equation}
where $\mathbf{H}_{i}^{hmt(est)} \in {\mathbb{R}^{w \times h}}$ is the heatmps corresponding to estimated joint coordinates and $\mathbf{H}_{i}^{hmt(gt)} \in {\mathbb{R}^{w \times h}}$ denotes heatmaps corresponding to ground truth joint coordinates.

\subsection{Baseline Architecture for Ablation Study}
As in Fig.\ref{fig_base_base}, this is the baseline architecture for ablation study. Heatmap guided \cite{du2019crossinfonet} is adopted as initial feature extraction module of this baseline. Besides, hand pose regression module is simple two cascaded residual modules that following fully-connected layer to estimate the final $3\times$J joints, which J is the number of estimated joints.

\subsection{Implementation Details}

The proposed network is implemented as end-to-end training manner using TensorFlow framework. The residual module is based on the public code in \cite{newell2016stacked}. The training process takes about $8$ hours on NYU and MSRA dataset, and $15$ hours on full ICVL dataset. All experiments were conducted on a Titan Xp with CUDA $9.0$ and cudnn $7.5$. In these experiments, batch size is set to $64$, and totally $110$ epochs in the training stage. The initial learning rate is set to $0.002$ and drop to $0.96$ for every epoch.

\section{Experimental Results}
In this chapter, we introduce three popular datasets for hand pose estimation to evaluate the performance of our proposed method, and describes the evaluation metrics of 3D hand pose estimation. In addition, ablation experiments are performed to verify the initial feature extracted module and hand morphological topology network structure. Comparison with the state-of-the-art methods is also conducted on NYU, ICVL and MSRA dataset. Qualitative results tested on these datasets are also shown in this section.

\subsection{Benchmarks}

In order to evaluate the performance of our method, we take sufficient experiments on three popular datasets, these are, ICVL, NYU, and MSRA dataset.

\subsubsection{ICVL Dataset}

The ICVL dataset \cite{tang2014latent} is from Imperial College. Descriptions of the dataset collected using Intel Creative Time-of-flight depth camera is as follows: Each line of the tag is a plot of coordinates, each line containing $16 \times 3$ dimensional data, representing $16$ joints center $(x,y,z)$ coordinates. $(x,y)$ is in pixels and $(z)$ is based on the true position, in (mm) unit. It also gives the order of the finger joints and the calibration parameters used to obtain the data set using the sensor.

\subsubsection{NYU Dataset}

The NewYork University Hand Pose Dataset \cite{tompson2014real} uses depth sensors to acquire images from two perspectives. The training set was collected from one person and the test set was obtained from two people. The image of the data set contains 36 key joints, $14$ of them are selected to hand pose estimation.

\subsubsection{MSRA Dataset}
MSRA dataset \cite{sun2015cascaded} come from Microsoft Research Asia, and the dataset contains 76500 frames from nine different subjects by SR300 depth camera. $4$ joints for each finger and $1$ joint for the palm, then $21$ joints were annotated. The dataset experiences large range of viewpoint changing, so it is very difficult to estimate the pose. In this experiment, we also compare the angle between the predicted and ground truth joint position.

%

\subsection{Evaluation Metrics}
In this paper, commonly used two metrics are engaged to evaluate the performance of hand pose estimation methods. As we used in the following part, these two metrics are average 3D distance error and success rate, respectively.

\begin{table}[tp]  
  \centering  
  \fontsize{8}{10}\selectfont  
  \caption{Comparsion of ablation study on ICVL, NYU, and MSRA dataset. In methods, The first "base" or "concat" means whether to use concatenated feature and the second "base" or "hbt" denotes whether to use hand morphological topology information. The evaluate error is in mm.}  
  \label{tab_ablation}  
    \begin{tabular}{c||ccccc}  
    \hline  
    \hline 
Methods & ICVL(22K) & NYU & MSRA(P0) \cr  
    \hline  
    \hline 
base+base           & 7.61   & 10.46 & 7.59 \\
concat+base         & 7.47   & 10.32 & 7.55 \\
base+hbt 		    & 6.89   & 9.12  & 7.34 \\
\textbf{concat+hbt} & 6.58   & 8.88  & 7.26 \\
    \hline  
    \hline 
    \end{tabular}  
\end{table}

\begin{figure*}[!t]
	\centering
	\subfigure{\includegraphics[width=2.3in]{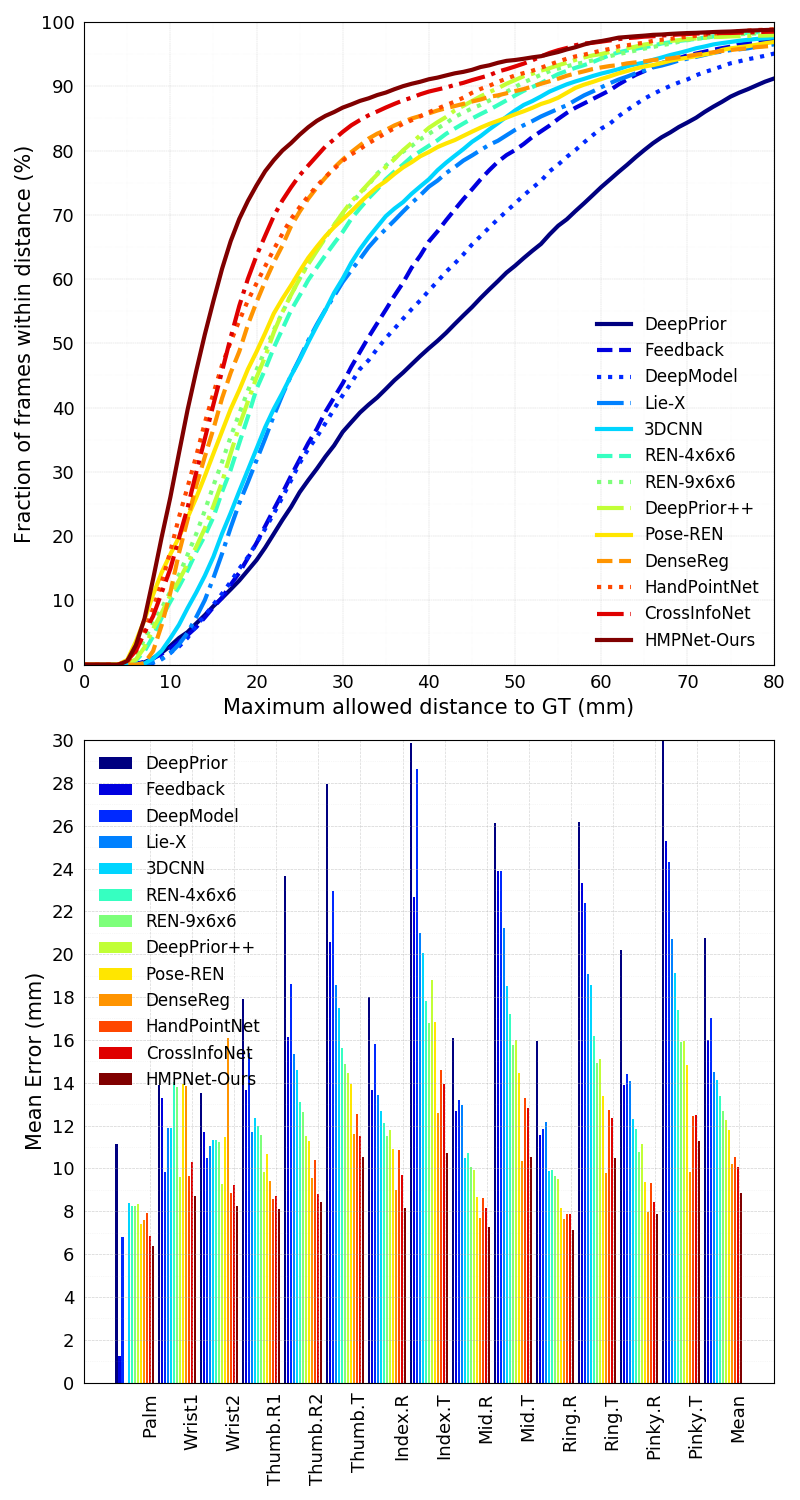}}
    \subfigure{\includegraphics[width=2.3in]{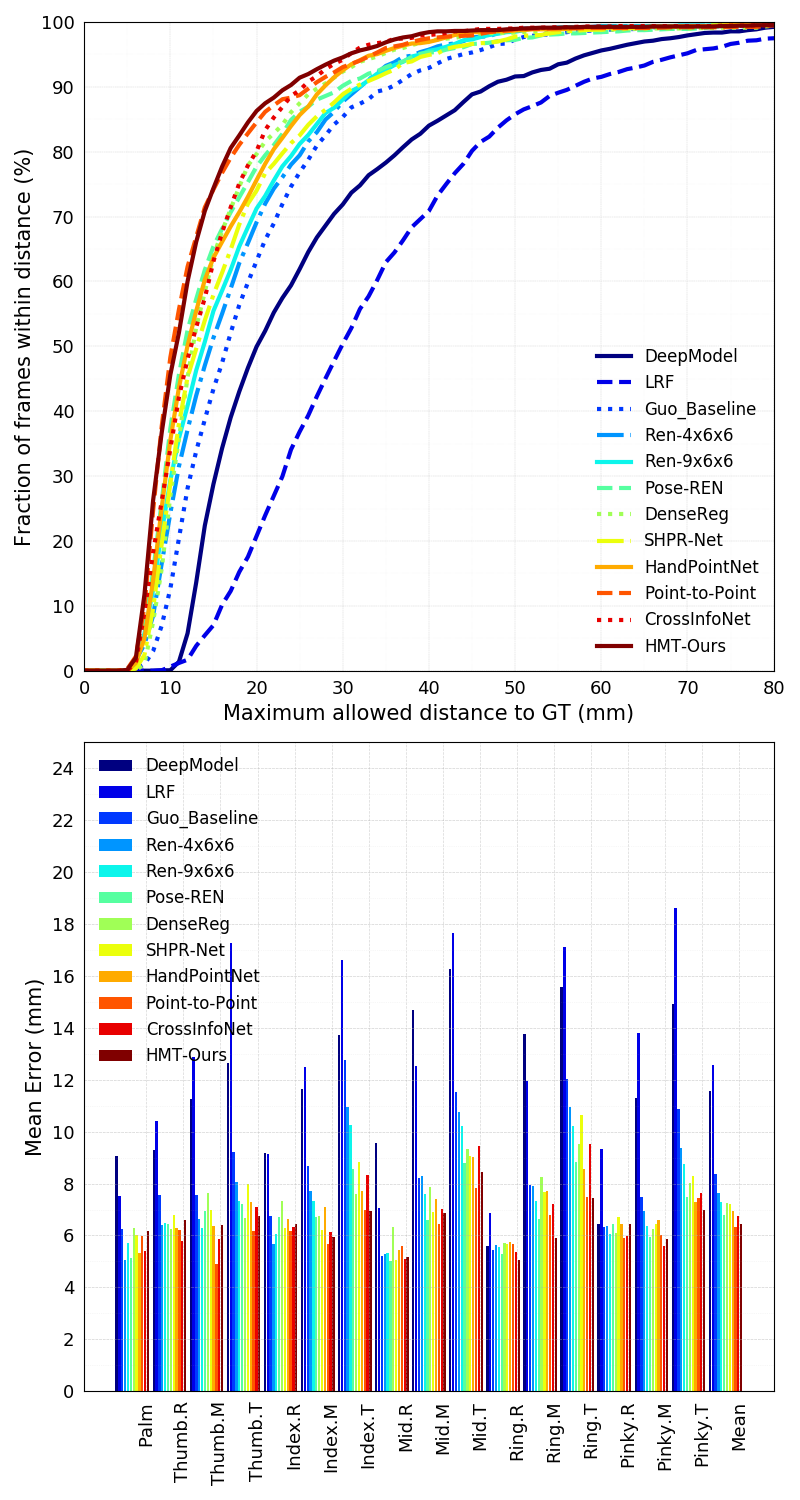}}
    \subfigure{\includegraphics[width=2.3in]{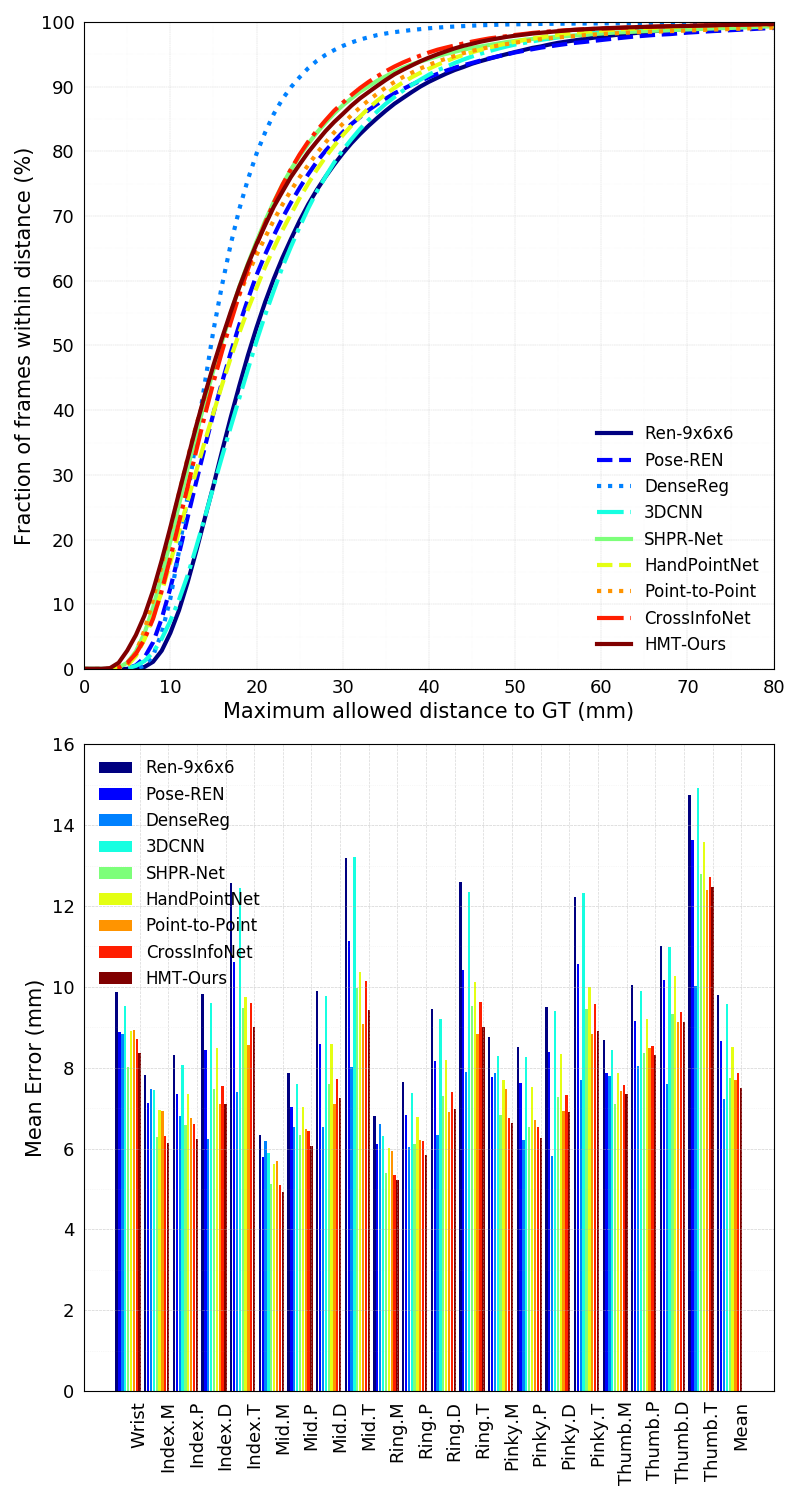}}
	
	\caption{Comparisons with state-of-the-art methods. Left: NYU dataset, Middle: ICVL dataset, Right: MSRA dataset}
	\label{fig_quativative}
\end{figure*}

\subsection{Ablation Study}

In order to verify our proposed method, sufficient ablation experiments are conducted on all three datasets as shown in Table \ref{tab_ablation}. To evaluate performance of our proposed network, we compare results of our network with that of baseline architecture without using concatenated feature and hand morphological structure. In these experiments, $22$K true samples from ICVL dataset are used for training and evaluation, and we only choose first subsection of MSRA dataset (P$0$) to evaluate our method.

\begin{table}[!t]  
  \centering  
  \fontsize{8}{10}\selectfont  
  \caption{Comparison with state-of-the-art methods on ICVL, NYU, and MSRA hand pose estimation datasets. fps represents frame per frame, which is inference time of our network architecture tested on a single GPU. Upper part is 3D based methods and below is 2D based methods.}  
  \label{tab_sota} 
    \begin{tabular}{c||ccc||c}  
    \hline  
    \hline 
Methods & ICVL & NYU & MSRA & fps \cr  
    \hline  
    \hline 
SHPR-Net \cite{chen2018shpr}  & 7.22 & 10.78 & 7.76 & - \\
3DCNN \cite{ge20173d} & -- & 14.1 & 9.6 & - \\
HandPointNet \cite{ge2018hand}    & 6.94 & 10.54 & 8.5 & - \\
Point-to-Point \cite{ge2018point} & 6.3  & 9.1   & 7.7 & 41.8 \\
V2V \cite{moon2017v2v}            & 6.28 & 8.42  & 7.59 & 3.5 \\
\hline
\hline
DeepModel\cite{zhou2016model}  & 11.56 & 17.04 & -- & - \\
DeepPrior\cite{oberweger2015hands}    & 10.4 & 19.73 & -- & - \\
DeepPrior++ \cite{oberweger2017deepprior++}   & 8.1 & 12.24 & 9.5 & - \\
Feedback \cite{oberweger2015training} & -- & -- & 15.97 & - \\
Lie-X \cite{xu2017lie}   & -- & -- & 14.51 & - \\
LRF\cite{tang2014latent}      & 12.58 & -- & -- & - \\
Ren-4x6x6 \cite{guo2017region}  & 7.63 & 13.39 & -- & - \\
Ren-9x6x6 \cite{wang2018region}  & 7.31 & 12.69 & 9.7 & - \\
DenseReg \cite{wan2018dense}   & 7.3 & 10.2 & 7.2 & 27.8 \\
Pose-Ren \cite{chen2017pose}   & 6.79 & 11.81 & 8.65 & - \\
CrossInfoNet \cite{du2019crossinfonet} & 6.73 & 10.08 & 7.86  & 124.5 \\
\textbf{HMPNet(Ours)} & 6.38 & 8.88 & 7.49 & 220.7\\
    \hline  
    \hline 
    \end{tabular}  
\end{table}

\begin{table}[!t]  
  \centering  
  \fontsize{8}{10}\selectfont  
  \caption{Cross validation of $9$ subsections on MSRA dataset. Mean error represents the mean three-dimensional distance error.}  
  \label{tab_crossvalidation}  
    \begin{tabular}{ccccccccc}  
    \hline  
    \hline 
P0 & P1 & P2 & P3 & P4  \cr
7.26 & 8.32 & 7.63 & 9.23 & 7.62 \cr
\hline
\hline
P5  & P6   & P7  & P8 & \textbf{Mean} \cr  
6.27 & 8.05 & 5.41 & 7.65 & \textbf{7.49}\cr
    \hline  
    \hline 
    \end{tabular}  
\end{table} 

\subsubsection{Concatenated Feature}

On ICVL dataset, using concatenated feature can reduce the error from $7.61$ mm to $7.47$ mm without HMT, and also from $6.89$ mm to $6.58$ mm with HMT, which reveals that the concatenated feature can decrease the estimation error greatly. The same conclusion can be drawn on the other two datasets from the right two columns of results.

\subsubsection{Hand Morphological Topology}

Similarly, hand morphological structure can improve the performance from $7.61$ mm to $6.89$ mm without concatenated feature, and $7.47$ mm to $6.58$ mm with concatenated feature on ICVL datset. On NYU dataset, the error drop from $10.32$ mm to $8.88$ mm whit concatenated feature.

\subsection{Comparisons with State-of-the-art Methods}

To evaluate the performance of our proposed method, we compare our results with the state-of-the-art methods in this subsection. We will discuss results on three datasets, respectively. On ICVL dataset, we achieved the best performance among these 2D methods, whose error is $0.1$ mm larger than V2V \cite{moon2017v2v} method, and $0.08$ mm worse than Point-to-Point method. However, V2V\cite{moon2017v2v} belongs to 3D method which is very time consuming and only can run at $3.5$ fps speed. Then, on NYU dataset, our results ($8.84$ mm) also are the second only inferior to V2V\cite{moon2017v2v} method which achieves ($8.42$ mm) error. On MSRA dataset, nine sub datasets are evaluated independently to evaluate the performance as shown in Table \ref{tab_crossvalidation} and Fig.\ref{fig_quativative}. To summarize, our proposed method is an effective and accurate method using only a single depth image compared with popular 2D based and 3D based methods.

\begin{figure*}[!tp]
	\centering
	\subfigure{\includegraphics[width=0.8in]{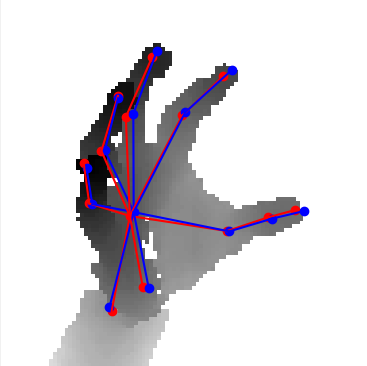}}
	\subfigure{\includegraphics[width=0.8in]{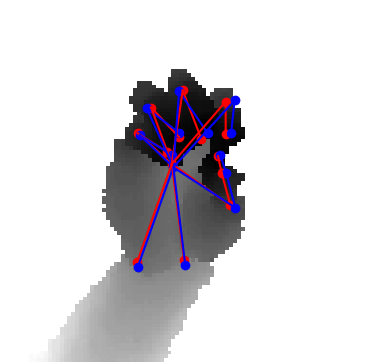}}
	\subfigure{\includegraphics[width=0.8in]{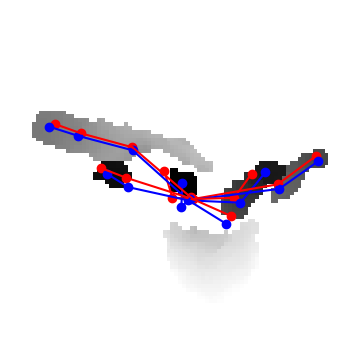}}
	\subfigure{\includegraphics[width=0.8in]{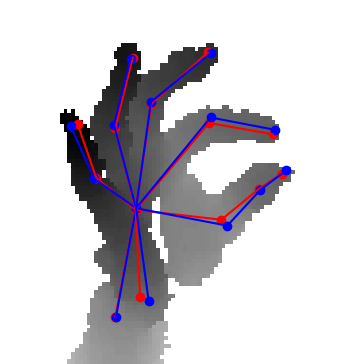}}
	\subfigure{\includegraphics[width=0.8in]{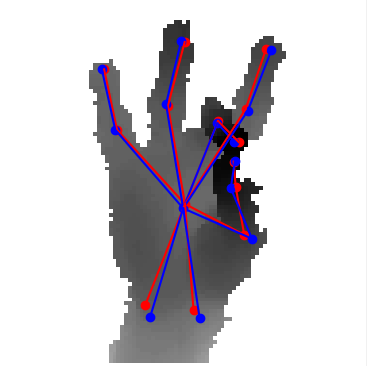}}
	\subfigure{\includegraphics[width=0.8in]{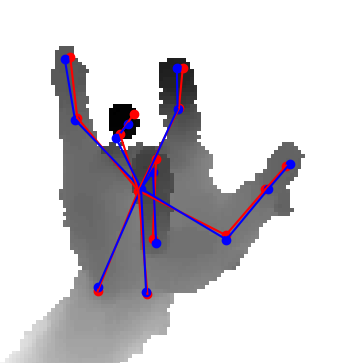}}
	\subfigure{\includegraphics[width=0.8in]{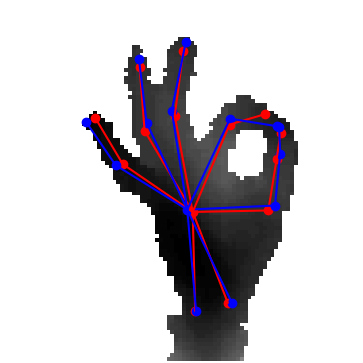}}
	\subfigure{\includegraphics[width=0.8in]{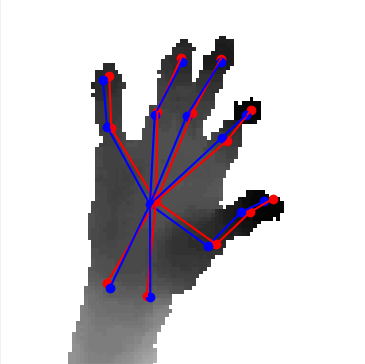}}

	\subfigure{\includegraphics[width=0.8in]{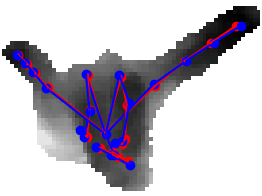}}
	\subfigure{\includegraphics[width=0.8in]{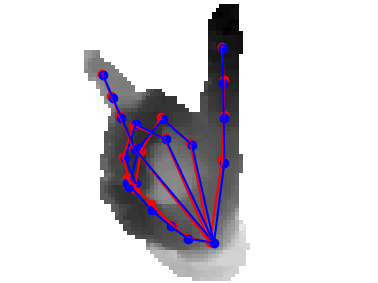}}
	\subfigure{\includegraphics[width=0.8in]{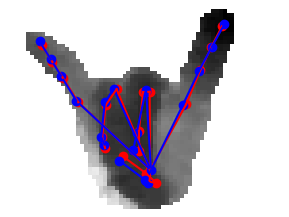}}
	\subfigure{\includegraphics[width=0.8in]{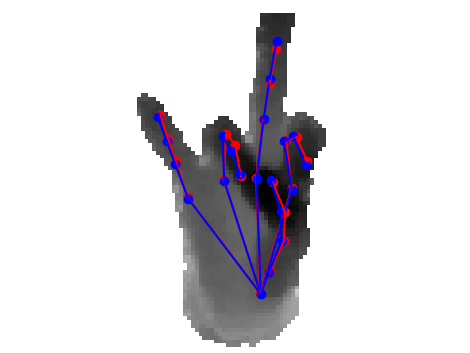}}
	\subfigure{\includegraphics[width=0.8in]{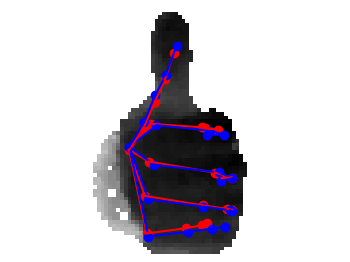}}
	\subfigure{\includegraphics[width=0.8in]{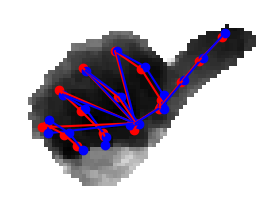}}
	\subfigure{\includegraphics[width=0.8in]{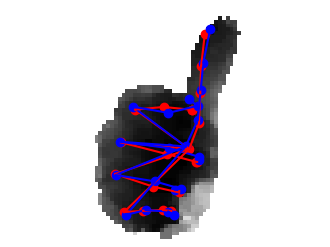}}
	\subfigure{\includegraphics[width=0.8in]{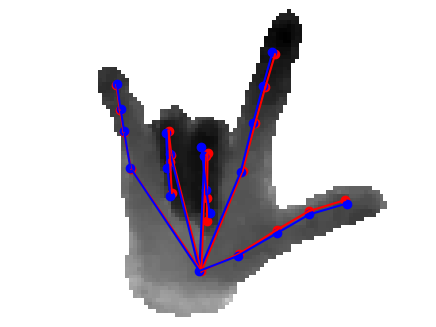}}
	
	\subfigure{\includegraphics[width=0.8in]{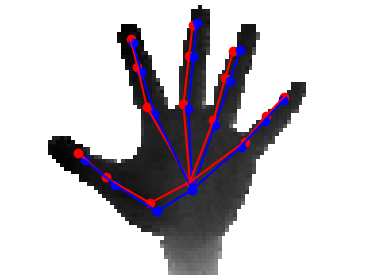}}
	\subfigure{\includegraphics[width=0.8in]{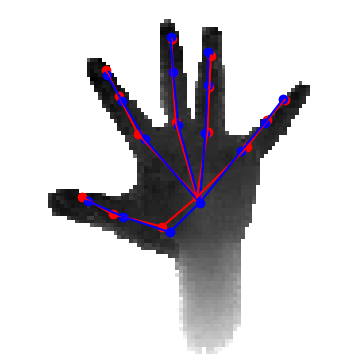}}
	\subfigure{\includegraphics[width=0.8in]{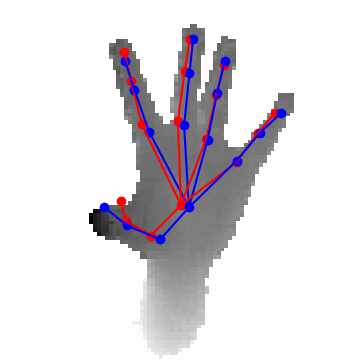}}
	\subfigure{\includegraphics[width=0.8in]{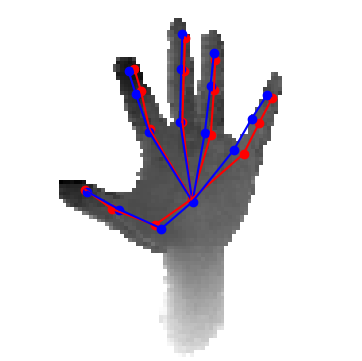}}
	\subfigure{\includegraphics[width=0.8in]{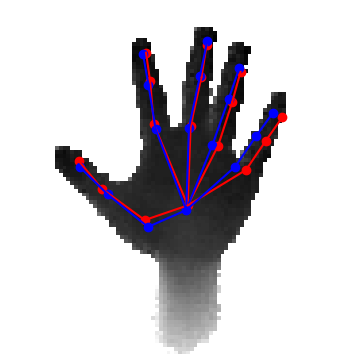}}
	\subfigure{\includegraphics[width=0.8in]{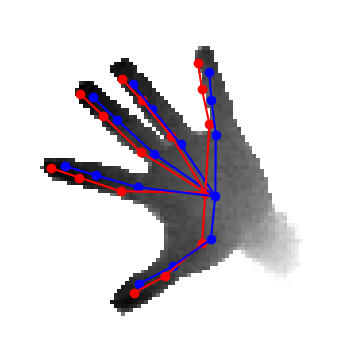}}
	\subfigure{\includegraphics[width=0.8in]{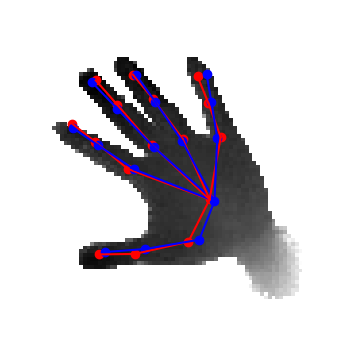}}
	\subfigure{\includegraphics[width=0.8in]{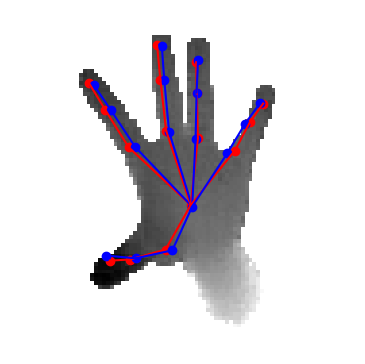}}
		
	\vspace{-0.1in}
	\caption{Qualitative results using our method on three hand pose estimation datasets. We compare our method with the ground truth joint locations. The predicted results are in red and the ground truth pose are in blue which are aligned on depth maps. These figures are better viewed in color. Top row: NYU dataset, Middle row: ICVL dataset, Bottom row: MSRA dataset.}
	\label{fig:handposequalitative}
\end{figure*}

\subsection{Runtime Analysis}
In realistic applications, running time of inference is very import. So we also compared our test time with other popular methods as shown in Table \ref{tab_sota}. In a practical application scenario, there may be more than one hand pose estimation task. Hence, this task should be more than $60$ fps speed. We can achieve about $220.7$ fps speed on a single GPU which meets the requirement of real-time applications. Although V2V \cite{moon2017v2v} and \cite{wan2018dense} achieved most accurate results, they only can run at $3.5$ fps and $27.8$ fps, respectively.

\section{Conclusions and Future Work}

In this paper, we focus on the 3D hand pose estimation task based on only a depth image. The main contributions are consisted of two aspects, one is a richer initial feature extraction module and the other is a tree-like structure network inspired from hand morphology topology. We derive heatmap corresponding to each joint according to the hand joint structure to learn the dependency between joints. This can be seen as the different range of motion via the distribution order belonging to each finger. Sufficient ablation experiments are implemented to verify these two components. In addition, we also obtained impressive results on popular dataset compared with famous 2D based methods even with 3D based methods. The results shown our proposed method is an effective and accurate hand pose estimation method using only a single depth image. In the future, we will take more insight into joint 3D hand-object pose estimation task or human robot interaction task.

%

%

\bibliographystyle{IEEEtran}
\bibliography{IEEEabrv,ms}

\EOD

\end{document}